\documentclass{article}

\usepackage{arxiv}

\usepackage[utf8]{inputenc} 
\usepackage[T1]{fontenc}    
\usepackage{hyperref}       
\usepackage{url}            
\usepackage{booktabs}       
\usepackage{amsfonts}       
\usepackage{nicefrac}       
\usepackage{microtype}      
\usepackage{lipsum}		
\usepackage{graphicx}
\usepackage{natbib}
\usepackage{doi}
\usepackage{amsmath}

\usepackage{caption}
\usepackage{subcaption}
\usepackage{threeparttable}
\usepackage{calligra}

\newtheorem{corollary}{Corollary}
\newtheorem{theorem}{Theorem}
\newtheorem{definition}{Definition}
\newtheorem{lemma}{Lemma}
\newtheorem{assumption}{Assumption}

\DeclareMathOperator{\tr}{Tr}

\DeclareMathAlphabet{\mathcalligra}{T1}{calligra}{m}{n}
\DeclareFontShape{T1}{calligra}{m}{n}{<->s*[2.2]callig15}{}
\newcommand{\scripty}[1]{\ensuremath{\mathcalligra{#1}}}

\title{Transfer Learning with Random Coefficient Ridge Regression}

\author{{Hongzhe Zhang} \\
	Department of Biostatistics, Epidemiology and Informatics, Perelman School of Medicine\\
	University of Pennsylvania\\
	Pennsylvania, PA 19104 \\
	\texttt{hongzhez@upenn.edu} \\
	\And{Hongzhe Li} \\
	Department of Biostatistics, Epidemiology and Informatics, Perelman School of Medicine\\
	University of Pennsylvania\\
	Pennsylvania, PA 19104 \\
	\texttt{hongzhe@upenn.edu} \\}
\renewcommand{\headeright}{}
\renewcommand{\undertitle}{}
\renewcommand{\shorttitle}{}

\hypersetup{
pdftitle={A template for the arxiv style},
pdfsubject={q-bio.NC, q-bio.QM},
pdfauthor={David S.~Hippocampus, Elias D.~Striatum},
pdfkeywords={First keyword, Second keyword, More},
}
\begin{document}
\maketitle
\begin{abstract}
Ridge regression with random coefficients provides an important alternative to fixed  coefficients regression in high dimensional setting when the effects are expected to be small but not zeros. 
This paper considers estimation and prediction of random coefficient ridge regression in the setting of  transfer learning, where in addition to observations from the target model, source samples from different but possibly related regression models are available. The informativeness of the source model to the target model can be quantified by the correlation between the regression coefficients. This paper proposes two estimators of regression coefficients of the target model as the weighted sum of the ridge estimates of both target and source models, where the weights can be determined by minimizing the empirical estimation risk or prediction risk.  Using random matrix theory, the limiting values of the optimal weights are  derived under the setting when $p/n \rightarrow \gamma$, where $p$ is the number of the predictors and $n$ is the sample size, which leads to an explicit expression of the estimation or prediction risks. Simulations show that these limiting risks agree very well with the empirical risks. An application to predicting the polygenic risk scores for lipid traits shows such transfer learning methods lead to smaller prediction errors than the  single sample ridge regression or Lasso-based transfer learning. 
\end{abstract}

\section{Introduction}
Massive and diverse data sets are ubiquitous in modern applications, including those in genomics and medical decisions.  It is of significant interest to integrate different data sets to obtain more accurate parameter estimates or to make a more accurate prediction of an outcome.   Given a target problem to solve, transfer learning \citep{Torrey10} aims at transferring the knowledge from different but related samples or studies  to improve the learning performance of the target problem. In biomedical studies, some  clinical  or biological outcomes are hard to obtain due to ethical or cost issues, in which case transfer learning  can be leveraged to boost the prediction and estimation performance by effectively utilizing information from related studies.
Transfer learning has been applied to problems in medical and biological studies, including predictions of protein localization \citep{Mei11}, biological imaging diagnosis \citep{Shin16}, drug sensitivity prediction \citep{Turki17}, and integrative analysis of ``multi-omics'' data, see, for instance, \citet{Sun16}, \citet{Hu19}, and \citet{Wang19}. It has also been applied to  natural language processing \citep{Daume07} and recommendation systems \citep{Pan13} in machine learning.  

In high dimensional setting, \cite{li2021transfer} developed transfer learning methods for sparse high dimensional regressions and demonstrated that one can improve  predictions of gene expression levels using data across different tissues. Such sparse models work well when the true models are sparse and the sample sizes are large.  However, there are settings where sparse model assumption may not be valid.   In genetics,  estiamting the polygenetic risk scores (\textsc{prs}s) using genome-wide genotype data \citep{mak2017polygenic,poly2} is an active area of research.   Such  \textsc{prs}s can be used in risk stratification, or can be treated as risk factors in population health studies. However, due to very large number of genetic variants but relatively small sample sizes, building a \textsc{\textsc{prs}} model that can accurately predict the \textsc{\textsc{prs}} scores is challenging. High dimensional regression models have been increasingly used to build the \textsc{prs} models based on genome-wide association (GWAS) data \citep{prs-lasso,prs-ridge,prs-ridge2}. However, sparse regression methods such as Lasso do not perform well for \textsc{prs} prediction with the current sample sizes due to the fact that the  genetic variants that are only weakly correlated with the outcomes, which are common for most genetic variants, are not selected.  Ridge regression, which does not require selecting the genetic variants and the sparseness assupmtion, but can handle the linkage disequilibrium among the genetic variants, provides a viable method for \textsc{prs} prediction. 

In this paper, we consider the problem of transfer learning for ridge regressions for \textsc{prs} predictions. Specially, the target polygenetic risk model is characterized by regression coefficients $\beta_K$, 
\begin{equation}\label{target.model}
	\underbrace{y}_{\text{Target Trait $K$}}  = \beta_{K1} \underbrace{x_1}_{\text{SNP 1}}  + \beta_{K2} \underbrace{x_2}_{\text{SNP 2}}   + \cdots + \beta_{Kp}\underbrace{x_p}_{\text{SNP p}} + \epsilon_K
\end{equation}
where $\beta_K=(\beta_{Kj}, j=1, \cdots, p)$ is the regression coefficients, and $x_j$ is the genotype score of the $j$th single nucleotide polymorphism (\textsc{snp}).   In the setting of ridge transfer learning, we assume that we additionally have data from $K-1$ source models given as
\begin{equation} \label{source.model}
	\underbrace{y}_{\text{Source Trait $k$}}  = \ \beta_{k1} \underbrace{x_1}_{\text{SNP 1}}  + \beta_{k2} \underbrace{x_2}_{\text{SNP 2}}   + \cdots + \beta_{kp}\underbrace{x_p}_{\text{SNP p}} + \epsilon_k,
\end{equation}
for  $k=1,\cdots, K-1$, where $\beta_k=(\beta_{kj}, j=1, \cdots, p)$ is the corresponding random coefficients for the $k$th model.  We particularly consider the  random regression coefficient (\textsc{rrc}) assumptions for both the target model \eqref{target.model} and  source models \eqref{source.model}. Specifically, we assume 
\begin{assumption}[Random Coefficientss Regression]
	\label{aspRCR}
\[Var(\epsilon_k) = \sigma_k^2, E({\beta_k}) = 0, Var[\sqrt{p} {\beta_k}] = \alpha_k^2 \sigma_k^2 I_p, \text{ for }  k = 1, \cdots, K\]
\[X_k \sim Z^k \Sigma^{1/2}, k=1,\cdots,K \mbox{ with } E[Z^k_{ij}]=0, Var[Z^k_{ij}]=1.\]
\end{assumption}
Here $\alpha_k^2$ is the ratio of the variance of linear coefficient $\beta_k$ and the residual variance $\sigma^2$, and it plays the role of a signal-to-noise ratio (\textsc{snr}). This  parameter is closely linked with genetic heratibility \citep{herit1}. When $X_k$ is properly scaled, the genetic heritability can be defined as 
\[h^2_k := \frac{\beta^T_k \Sigma \beta_k}{\beta^T_k \Sigma \beta_k + \sigma_{k}^2} \rightarrow_{a.s.} \frac{\alpha_k^2 \sigma_k^2}{\alpha_k^2 \sigma_k^2 + \sigma_k^2} = \frac{\alpha_k^2}{1 + \alpha_k^2}\]
Standard consistent estimators exist for both $h^2_k$ and $\sigma_k^2$ \citep{lee2012estimation}, so $\alpha_k^2$ can also be reliably estimated. We  treat them as fixed and  known parameter in this paper.

Ridge-type regressions under the \textsc{rrc} assumptions have been studied by \citet{dobriban2018high}, \citet{sheng2020} and \citet{zhao2019cross}. \citet{dobriban2018high} was the first to use the random matrix theory (\textsc{rmt}) results on trace of functions of sample and population covariances to find the limiting prediction risk of ridge regression. \citet{sheng2020} extended  ridge regression to a distributed learning setting, and \citet{zhao2019cross} studied the in-sample and out-of-sample $R^2$ of several ridge-type estimators using the same \textsc{rmt} results. A difference is that their results are stated mostly in terms of moments of population spectral distributions. As another setup, \cite{richards2021asymptotics} assumes $\beta$ come from a structured prior distribution and the larger coefficients are more in line with the principal components of population covariance matrix. Similarly, \citet{GenRid1} also assume a non-isotropic prior for $\beta$ but they study the generalized ridge regression where the penalization on $\beta$ is also non-isotropic. They both focused on the asymptotic performance of the overparameterized models under the proportional regime. \citet{dimFree} and \citet{Trevor2022} treat $\beta$ as fixied and provided non-asymptotic bounds that approximate the prediction risk for ridge regressions. \citet{Trevor2022}  build the results on recent advances in random matrix theory \citep{knowles2017anisotropic} while \citet{dimFree} find the prediction risk of an ‘equivalent’ sequence of regression models with diagonal design matrix.  \citet{dimFree} generalizes the results in \citet{Trevor2022} by allowing the number of variables and obervations to grow non-proportionally, and its approximation error is multiplicative, thus the approximation is more accurate when the risk is small.

Among these works, \citet{zhao2019cross} and \citet{sheng2020} aim to  obtain a better estimator for $\beta$ by integrating information from multiple sources of data. Both papers  used a weighted sum of naive estimator as an aggregated etimator:
 \[\hat{\beta} =  \sum_{k = 1}^{K} W_k  \hat{\beta}_{k}, \]
where $\hat{\beta}_{k}$ is the estimator based on  data from the $k^{th}$ population \citep{sheng2020}. They study the aggregated ridge estimator that optimizes an estimation risk by assuming  a common set of linear coefficients $\beta$ in all populations. \citet{zhao2019cross} proposes an aggregated marginal estimator that ignores the correlation structure of the design matrix to optimize the out-of-sample $R^2$. Although they still assume a common $\beta$ in all training populations, their objective is to explain the most variance in an ''out-of-sample" population with different  coefficients. 

We consider a framework of transfer learning  by assuming the coefficients $\beta_k$ are  different in each of the training populations. More specifically, we assume all polygetic models are linked by the  correlations of  their random coefficients:
\begin{assumption}[Correlated Random Coefficients]
	\label{aspTL}
	\[ 
	Cov( \sqrt{p} {\beta_k}, \sqrt{p} \beta_{k'}) = \rho_{k k'} \alpha_k \alpha_{k'} \sigma_k \sigma_{k'} I_p \ \text{for} \ k, k' = 1, \cdots, K, k\neq k'.
	\]
\end{assumption}
Intuitively, the correlation coefficient $\rho_{kK}$ measures the similarity of random cofficients $\beta_k$ and $\beta_K$, therefore the amount of information available to target model in source $k$. Like $\alpha^2_k$ and $h^2_k$,  $\rho_{k k'}$ has a close connection to  genetic correlation \citep{herit1},  $\varphi_{k k'}$.   More specifically, the genetic correlation between two phenotypes  converges almost surely to $\rho_{k k'}$ when $n, p \rightarrow \infty$,
\[\varphi_{k k'} = \frac{\beta_k^T \Sigma \beta_{k'}}{\sqrt{\beta_k^T \Sigma \beta_{k} \beta_{k'}^T \Sigma \beta_{k'} }} \rightarrow_{a.s.} \frac{\rho_{k k'} \alpha_k \alpha_{k'} \sigma_k \sigma_{k'}}{\sqrt{\alpha_{k}^2 \sigma_{k}^2 \alpha_{k'}^2 \sigma_{k'}^2}} = \rho_{k k'}.\]
The parameters $\rho_{k  k'}$ can  be estimated reliably with the same method \citep{lee2012estimation}, and we  treat them as known fixed parameters.

In our transfer learning framework, we wish to optimize the performance of $\hat{\beta}$ for target $K^{th}$ population only. To determine the weight $W$ in the weighted estimate $\hat{\beta}$, we consider two different approaches,  the optimal limiting weight that minimizes the estimation risk, and  the optimal limiting weight that minimizes prediction risk. We show that  the genetic correlations between the target and source models determines  how much knowledge of a source model can be transferred to target model, and  study the performance of the aggregated estimator in a transfer learning set up. We provide the limiting prediction risks and estimation risks of the aggregated estimator that can be consistently estimated with empirical observations under general covariance matrix $\Sigma$ with no requirement on $n_k$.  The limiting risks rely  on evaluating the limits of the following two normalized trace terms
\begin{align*}
	&\tr\{\Sigma (\hat{\Sigma}_k + \lambda_k I_p)^{-1}  (\hat{\Sigma}_{k'} + \lambda_{k'} I_p)^{-1}\} / p, \\
	&\tr\{(\hat{\Sigma}_k + \lambda_k I_p)^{-1}  (\hat{\Sigma}_{k'} + \lambda_{k'} I_p)^{-1}\} / p,  
\end{align*}
where $\Sigma$ is the population covariance matrix, which is deterministic under our assumptionn,  and $\hat{\Sigma}_k := X^T_k X_k / n_k$ is the sample covariance matrix for population $k$. We derive these results under general distribution of $X$ using the anisotropic local laws \citep{knowles2017anisotropic} for random matrices.

\section{Estimation via transfer learning}
\subsection{Optimal Estimation Weights}
We assume that for each source  model and the target model, we have $n_k$ $i.i.d.$ observations, $(Y_k, X_k)$ ($k=1, \cdots, K$), where $Y_k$ is the vector of outcomes, and $X_k$ is  random $n_k \times p$ dimensional matrix of the genotype scores.  Denote the naive ridge estimator in source  $k, k = 1,\cdots, K$ as 
\[\hat{\beta}_{k} := (X_k^T X_k + n_k \lambda_k I_p)^{-1} X^T_k Y_k = (\hat{\Sigma}_k + \lambda_k I_p)^{-1} X^T_k Y_k / n_k, \]
where $\hat{\Sigma}_k : = X_k^T X_k/n_k$ is the sample covariance matrix for population $k$, and $\lambda_k$ is the tuning parameter.  We propose to estimate $\beta_{K}$ with a weighted sum of naive ridge estimators:
\begin{align}
	\hat{\beta} =  \sum_{k = 1}^{K} W_k \hat{\beta}_{k},
\end{align}
where $W_k$ is the weight for the $k^{th}$ source study. We firstly consider using the weights that minimize  the expected estimation risk as in \citet{sheng2020}.
\begin{definition}[Expected Estimation Risk] \label{estR}
	The expected estimation risk of an estimator $\hat{\theta}$ is defined as
	\[M_K(W, \lambda_1, \cdots, \lambda_{K}) = E_{\epsilon_1, \cdots, \epsilon_K}\left(||  \sum_{k = 1}^{K} W_k \hat{\beta}_{k} - \beta_K||^2 | X_1, \cdots, X_K\right),\]
\end{definition}
where  $W=(W_1, \cdots, W_K)^T$. We denote the estimation risk as $M_K(W)$ for the simplicity of notation, and  the optimal estimation weight that minimizes the $M_K(W)$ as $W_E^*$. The following theorem states the expressions for $W_E^*$ as well as $M_K(W_E^*)$.
\begin{theorem}[Finite Sample Optimal Weights and Estimation Risks] \label{optWeight}
Define the following quantities
\begin{align*}
	Q_k &:= (\hat{\Sigma}_K  + \lambda_K I_p)^{-1} \hat{\Sigma}_k \\
	B &:= [ Q_1 \beta_1 , \cdots,  Q_K \beta_K] \\
	v := B^T \beta_K &= vec[\beta_k^T Q_k \beta_K]  \\
	A := B^T B &= mat[\beta_k^T Q_k Q_{k'} \beta_{k'}] \\
	R &= diag[n_k^{-1} \sigma_k^2 \tr\{(\hat{\Sigma}_k + \lambda_k I_p)^{-2} \hat{\Sigma}_k\}]
\end{align*}
The optimal weights and the corresponding optimal risk are
\begin{eqnarray}W^*_E &=& (B^T B + R)^{-1} B^T \beta_K,\\
M_K(W^*_E)& =& \beta_K^T \{I_p - B(B^TB + R)^{-1} B^T\} \beta_K.
\end{eqnarray}
\end{theorem}
The results in Theorem 1 holds under finite samples, but as we shall see, their limiting forms, which depend on the unknown linear coefficients $\beta_k$ only through $\alpha^2_k$ and $\sigma_{k}^2$,  
 are more useful. 
\subsection{\textsc{rmt} background}
Our derivation of the limiting values of the weights and estimation risk is based on results of random matrix theory (\textsc{rmt}).  Following \citet{dobriban2018high},  \citet{sheng2020},  \citet{zhao2019cross}, we consider the Marchenko-Pastur  type sample covariance matrices. Define $F_{A,p}$ as the spectral distribution for a $p\times p$ symmetric matrix $A$ that places equal point mass on the eigenvalues $\lambda_i(A)$ of $A$ with the corresponding cumulative distribution function (CDF)  defined as $F_{A,p}(x) := \sum_{i = 1}^p  \mathbb{1}(\lambda_i(A) \leq x)/p$, where $\mathbb{1}(\cdot)$ is the indicator function. Our analysis relies on the convergences of eigenvalue distributions of sample covariance matrices, which require the following assumptions.

\begin{assumption}[\textsc{rmt} assumption]\label{aspRMT} \
	\begin{enumerate}
		\item  For $k = 1, \cdots, K$, the $n_k \times p$ design matrix $X_k$ is generated as $X_k = Z_k \Sigma^{1/2}$ for an $n_k \times p$ matrix $Z_k$ with $i.i.d.$ entries coming from an infinite array. The entries $(Z_k)_{ij}$ of $Z_k$ satisfying $E[(Z_k)_{ij}] = 0$ and $E[(Z_k)_{ij}^2] = 1$. We assume the $p \times p$ population covariance matrix is deterministic and positive semidefinite.
		\item The sample size $n_k$ grows to infinity proportionally with the dimension $p_k$ for $k = 1, \cdots, K$, i.e. $min(n_1, p_1, \cdots, n_K, p_K) \rightarrow \infty$ and $p_k / n_k \rightarrow \gamma_k \in (0, \infty)$.
		\item The sequence of spectral distributions $F_\Sigma := F_{\Sigma, n, p}$ of $\Sigma := \Sigma_{n,p}$ converges weakly to a limiting distribution $H$ supported on $[0, \infty)$, called the population spectral distribution (\textsc{psd}).
	\end{enumerate}
\end{assumption}

Given the \textit{RMT} assumption, the Marchenko-Pastur theorem then states that with probability one, the empirical spectral distribution (ESD) $F_{\hat{\Sigma}}$ of the sample covariance matrix $\hat{\Sigma}$ also converges weakly (in distribution) to a limiting distribution $F_{\gamma} := F_{\gamma}(H)$  supported on $[0, \infty)$. The ESD of sample covariance matrix is determined uniquely by a fixed-point equation for its \textit{Stieltjes transform}, which is  defined  as
\[m_G(z) = \int_{l = 0}^\infty \frac{d G(l)}{l-z}, \ z \in C \ \backslash \ R^+, \]
for any distribution $G$ supported on $[0, \infty)$. Given this notation, the Stieltjes transform of the spectral measure of $\hat{\Sigma}$ satisfies
\[m_{\hat{\Sigma}}(z) = \tr\{(\hat{\Sigma}- z I_p)^{-1}/p\} \rightarrow_{a.s.} m_F(z). \]
We  define the companion Stieltjes transform of the limiting spectral distribution of the $n \times n$ matrix ${\hat{\Sigma}}  = XX^T/n$ as $v(z)$. For all $z \in C \ \backslash \ R^+$, the Stieltjes transform $v_F(z)$ is related to $m_F(z)$ by
\[\gamma [m_F(z) + \frac{1}{z}] = v_F(z) + \frac{1}{z}.\]
In addition, we denote by $m'_F(-\lambda)$ the derivative of the Stieltjes transform $m_F(z)$ evaluated at $z = -\lambda$, where 
\[m'_F(z) = \int_{l = 0}^\infty \frac{d G(l)}{(l-z)^2} \ \ \ v'_F(z) = \gamma (m'_F(z) - \frac{1}{z^2}) + \frac{1}{z^2}. \]
In terms of the empirical quantities,
\[\tr\{(\hat{\Sigma}- z I_p)^{-2}/p\} \rightarrow_{a.s.} m'_F(z). \]
\subsection{Asymptotic behavior when $p, n \rightarrow \infty$}
As in \citet{sheng2020}, we  provide the limiting values for optimal weights and estimation risks in terms of  the limiting ESD $F_\gamma$ of $\hat{\Sigma}_k$ under the setting that the aspect ratio $p_k \rightarrow \infty, n_k \rightarrow \infty, p_k/ n_k \rightarrow \gamma_k, 0<\gamma_k< \infty, k = 1 , \cdots, K$. We present the results only in terms of the empirical quantities which we can be easily estimated from the observations. In particular, our results do not involve the  \textsc{psd} $H$ of $\Sigma$, although parallel results can be easily derived. Most of the terms in the limits of estimation (Theorem \ref{estAsp}) and prediction risks (Theorem \ref{predOptAsp}) can be easily expressed given the prior work \citep{dobriban2018high}; some off-diagonal terms in matrix $\mathcal{A}$ and $\mathcal{C}$ involve the traces of products of sample covariance matrices from different studies. More explicitly, we are interested in the limit of the terms
\[E_{k k'} := \tr \{ (\hat{\Sigma}_k + \lambda_k I_p)^{-1} (\hat{\Sigma}_{k'} + \lambda_{k'} I_p)^{-1}/p\}, \] 
\[P_{k k'} := \tr \{\Sigma (\hat{\Sigma}_k + \lambda_k I_p)^{-1} (\hat{\Sigma}_{k'} + \lambda_{k'} I_p)^{-1}/p\}, \] 
which are derived  separately and presented as  Lemma \ref{aijasp} and Lemma \ref{predijasp} below. The following assumption \ref{assmoment} and assumption \ref{assIso} are needed when the number of observations or penalization parameters are different across different studies.
\begin{assumption}[Bounded Moment] \label{assmoment}
	Assume for each natural number $p$, the entries of $Z$ written by $Z_{ij}$ has uniformly bounded $p$-th moment. That is, there are constants $C_p$ such that
	\[E|Z_{ij}|^p \leq C_p\]
\end{assumption}
For general covariance matrices, we   need the following assumption to make sure the spectrum of $\Sigma$ is not concentrated around zero.
\begin{assumption}[Anisotropic Local Laws] \label{assIso}
	Denote the eigenvalues of $\Sigma$ by
	\[\sigma_1 \geq \sigma_2 \geq \cdots \geq \sigma_p \geq 0\]
	and let $\pi$ denote the empirical spectral density of $\Sigma$
	\[\pi := \frac{1}{p} \sum_{i = 1}^{p} \delta_{\sigma_i} \]
	For a positive constant $\tau >0$ that can be arbitrarily small, assume 
	\[\pi ([0, \tau]) \leq 1 - \tau\]
\end{assumption}
The following Lemma  summarizes the limiting value of $E_{k k'}, k \neq k' ; k, k' = 1,\cdots, K$ under different assumptions. The results for $P_{k k'}$ will be presented later.
\begin{lemma}\label{aijasp}
	Assume $n_1, \cdots, n_K, p \rightarrow \infty$, $p / n_k \rightarrow \gamma_k$ for $k = 1, \cdots, K$ and assumption \ref{aspRMT}, also use a shorter notation $m_k(-\lambda_k) := m_{F_{\gamma_k}} (-\lambda_k), m'_k(-\lambda_k) := m'_{F_{\gamma_k}} (-\lambda_k)$. We have 
	\[E_{k k'} \rightarrow_{a.s.} \mathcal{E}_{k k'}\]
	(1).
	 Assume $n_1 = \cdots = n_K = n$, so $\gamma_1 = \cdots = \gamma_K = \gamma$, and use $\lambda_1 = \cdots = \lambda_K = \lambda$.  Then by definition $m_k(-\lambda) = m_{k'}(-\lambda) := m(-\lambda)$ and $m_k'(-\lambda) = m_{k'}'(-\lambda) := m'(-\lambda)$. We have for $k \neq k'$
	\[\mathcal{E}_{k k'} = \frac{(1 - \gamma) m'(-\lambda) + 2 \gamma \lambda m(-\lambda) m'(-\lambda) - \gamma m(-\lambda)^2}{1 - \gamma +\gamma \lambda^2 m'(-\lambda)}. \]\\
	(2). 
	Under  the assumption $\Sigma = I_p$,
	\[\mathcal{E}_{k k'} = m_k(-\lambda_k)  m_{k'}(-\lambda_{k‘})\]
	(3). 
	Under the Assumptions \ref{assmoment} and \ref{assIso}, we have
	\begin{eqnarray*}
	\mathcal{E}_{k k'}& = & \frac{1}{\lambda_k \lambda_{k'}} \left\{ \lambda_{k} m_k(-\lambda_k) + \lambda_{k'} m_{k'}(-\lambda_{k'})  
	+ \frac{\lambda_k m_k(-\lambda_k) m_{k'}(-\lambda_{k'})}{ (m_k(-\lambda_k) - m_{k'}(-\lambda_{k'}))} \right. \\
	&& -\left.  \frac{\lambda_{k'} m_k(-\lambda_k) m_{k'}(-\lambda_{k'})}{ (m_k(-\lambda_k) - m_{k'}(-\lambda_{k'}))} \right\}.
	\end{eqnarray*}
\end{lemma}

We comment briefly on the proof for the three statements. The second case is the easiest and follows directly from asymptotic freeness between covariance matrices under isotropic population covariance matrix. In the first case, when the sample sizes are the same across all studies and the penalization parameters are the same, all studies,  $1, \cdots, K$, share the same unique solution: $x := x_{H, \gamma, \lambda}$ to a fixed point equation \citep{GMP}:
\[1 - x = \gamma \left\{1 - \lambda \int \frac{1}{x t + \lambda} dH(t)\right\} = \gamma [1 - \lambda m_{F_\gamma} (-\lambda)],\]
and the fact
\[\mathcal{A}_{k k'} = \rho_{k k'} \sigma_k \sigma_{k'} \alpha_k \alpha_{k'}\left\{\int \frac{x^2 t^2}{(xt +\lambda)^2}dH(t)\right\}.\]
The solution $x$ and its derivative w.r.t. $\lambda$ connect the unknown integral in $\mathcal{A}_{k k'}$ with empirical quantities $m_{F_\gamma} (-\lambda)$ and $m'_{F_\gamma} (-\lambda)$, which  can estimated  consistently  given Assumption \ref{aspRMT}. In the most general third case, no such common solution exists, and we apply  the anisotropic local laws \citep{knowles2017anisotropic}
\[\tr \{(\hat{\Sigma} + \lambda I_p)^{-1}\} \approx \frac{1}{\lambda} \tr \{(I_p + m(-\lambda) \Sigma)^{-1}\}.\]
This approximation  helps us to evaluate the integral by breaking it into functions of $m_{F_\gamma} (-\lambda)$. 

Using Lemma \ref{aijasp}, we can express the limiting optimal estimation weight and limiting estimation risk, as stated in the following theorem.

\begin{theorem}[Asymptotic Optimal Weights and Estimation Risk] \label{estAsp} Under the Assumptions \ref{aspRCR},  \ref{aspTL} and   \ref{aspRMT}. Assume further that the entries of $Z$ have a finite $(8 + \epsilon)$-th moment for some $\epsilon > 0$, and all eigenvalues of $\Sigma$ are uniformly bounded away from zero and infinity. We have 
	\begin{enumerate}
		\item $v \rightarrow_{a.s.} V$, where 
		$V_k = \rho_{k K} \sigma_k \sigma_K \alpha_k \alpha_K \{1 - \lambda_k m_{F_{\gamma_k}} (-\lambda_k)\}$ for $k = 1, \cdots K-1$, and 
		$V_K = \sigma_K^2 \alpha_K^2 \{1 - \lambda_K m_{F_{\gamma_K}} (-\lambda_K)\}.$
		\item $A \rightarrow_{a.s.} \mathcal{A}$, where for $k = 1, \cdots, K$, 
	$\mathcal{A}_{kk} = \sigma_k^2  \alpha_k^2 \{ 1 - 2 \lambda_k m_{F_{\gamma_k}} (-\lambda_k) + \lambda_k^2 m'_{F_{\gamma_k}}(-\lambda_k)\}$
	and 
$
		\mathcal{A}_{k k'} = \rho_{k k'} \sigma_k \sigma_{k'} \alpha_k \alpha_{k'} \{ 1 -  \lambda_k m_{F_{\gamma_k}} (-\lambda_k) - \lambda_{k'} m_{F_{\gamma_{k'}}} (-\lambda_{k'}) + \lambda_k \lambda_{k'} \mathcal{E}_{k k'}\}
	$
for $ k, k'  = 1, \cdots, K; k \neq k'$.
		\item Recall $R$ is a diagonal matrix, and $R \rightarrow_{a.s.} \mathcal{R}$, where   
		\begin{align*}
			\mathcal{R}_{kk} = \sigma_k^2 \gamma_k \{m_{F_{\gamma_k}} (-\lambda_k) - \lambda_k m'_{F_{\gamma_k}} (-\lambda_k)\},
			k = 1, \cdots, K.
		\end{align*}
	\end{enumerate}
	The limiting estimation weight is
	$\mathcal{W}^*_E =(\mathcal{A} + \mathcal{R})^{-1} V $, and the limiting estimation risk under the optimal weights is 
$\mathcal{M}^*_K(\mathcal{W}^*_E) = \alpha^2 \sigma_K^2 -  V^T (\mathcal{A} + \mathcal{R})^{-1} V. $
\end{theorem}
\begin{figure}[h]
     \centering
     \begin{subfigure}[b]{0.6\textwidth}
         \centering
         \includegraphics[width=\textwidth]{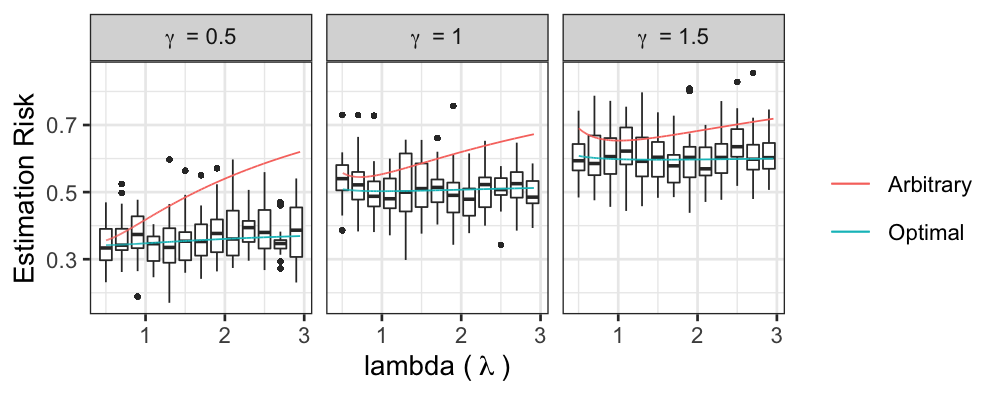}
     \end{subfigure}
        \\[\smallskipamount]
     \begin{subfigure}[b]{0.6\textwidth}
         \centering
         \includegraphics[width=\textwidth]{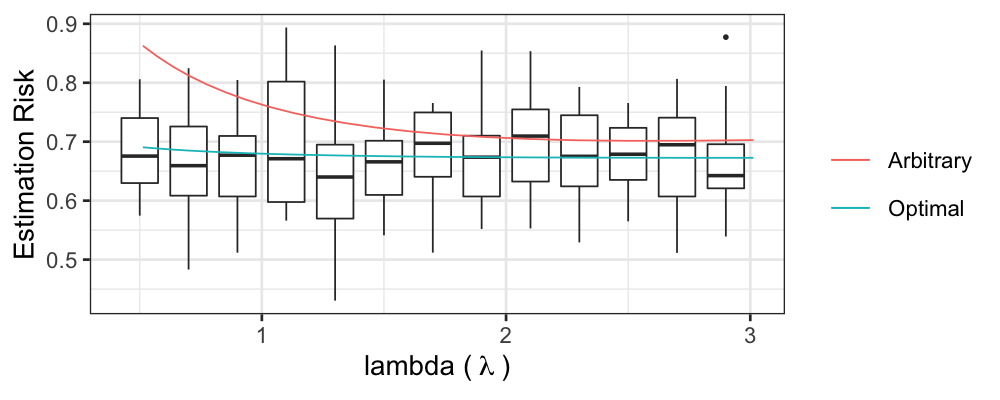}
     \end{subfigure}
     \\
     \begin{subfigure}[b]{0.6\textwidth}
         \centering
         \includegraphics[width=\textwidth]{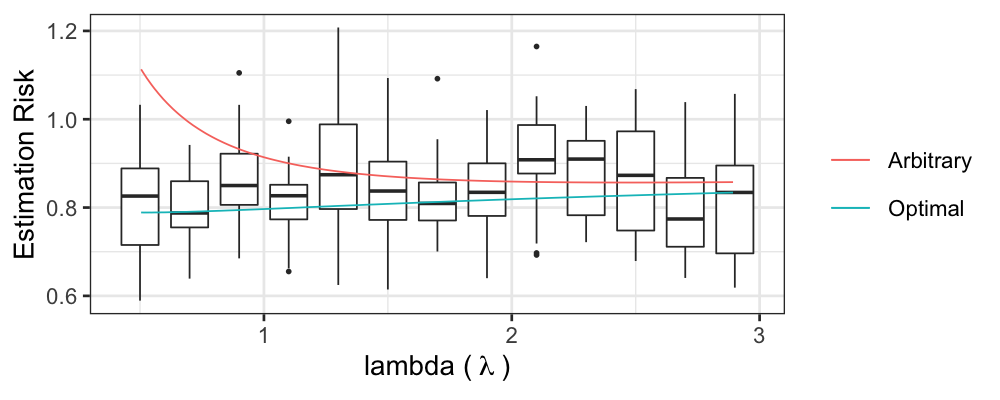}
         \label{fig:five over x}
     \end{subfigure}
        \caption{Empirical estimation risk and theoretical risk, where $\lambda$ is the ridge tuning parameter and $\gamma=n/p$.  The colors of two solid lines represent the two different weights.  Top: All studies have the same number of observations. When $\gamma$ increases, the improvement from using the optimal weights increases, compared with the optimally tuned arbitrary weights.
	Middle:  theoretical estimation risk under identity covariance matrix with unequal sample sizes.  Bottom:  unequal sample sizes and general covariance matrix.}
	\label{EstTheo_Sigma}
\end{figure}
We compare the theoretical estimation risks with simulated results in Figure \ref{EstTheo_Sigma}. We consider three different simulation setups, each corresponding to a set up in Lemma \ref{aijasp}, 
 where we keep the number of predictors  at  $p=100$ in all setups for all studies, and the genetic correlations between all the studies   as  $\rho=0.5$. In the top plot, the number of observations, $100 \times \gamma$, is the same for target and five source studies, but we vary $\gamma$ from $0.5$ to $1.5$. For the middle and the bottom plots, we set the number of observations in all six studies to decrease  from $150$ to $50$, with the target study with 50 observations. In addition, the population covariance matrices of the top and the bottom plots share a Toeplitz structure, while the middle plot has an identity covariance matrix. The asymptotics are very accurate in all three scenarios across a wide range of $\lambda$ values.  Our results are not very sensitive to the choice of the tuning  parameters.

To demonstrate the effectiveness of using the optimal weights ($\mathcal{M}^*_K(\mathcal{W}^*_E)$), in all three setups, we show  estimation risk $\mathcal{M}^*_K(\mathcal{W}^A)$ using equal weights  $\mathcal{W}_A$, 
$W^A_K = 1, W^A_1 = \cdots = W^A_{K-1} = {1}/{K}.$
As expected, the equal  weight risk is dominated by the optimal weights in estimation risks.  When the $\gamma_k$ are different, the optimal weights can adapt to study sizes, where larger study gets more weight, and therefore obtains significantly lower estimation risks. When $\gamma$ is larger, the gap between red ang blue lines gets larger. This suggests that when the available information gets more scarce,  choosing the  weights optimally  becomes more important. 

The genetic correlation also plays a crucial role determining the weights assigned to  target and source studies. We show in the next section that $\mathcal{W}^*_E$ has the expected adaptive properties. For the proposed estimator, however, we cannot provide a theoretical optimal $\lambda$. Such results are possible for the weighted estimator  under more restricted setup in \citet{sheng2020}, where the  populations are "exchangeable", thus the total estimation risk can be decomposed to the sum of naive ridge estimation risk in each study. One can then choose the $\lambda_{k}$ as the tuning parameter that minimizes naive estimation risks \citep{dobriban2018high}. In our setting, however, such a decomposition is impossible, as the prediction performance is evaluated only on target population $K$.  For real data, we recommend tuning the $\lambda_{k}$  using  cross-validations.

\section{Adaptivity of $\mathcal{W}^*_E$ to the informativeness of the source data}
To show the adaptive property of the weight vector $\mathcal{W}^*_E$, we  present  Corollary \eqref{Extrem} on  the limiting behavior of $\mathcal{W}^*_E$ when all $\rho_{kk'}, k,k' \in \{1, \cdots, K\}, k \neq k'$ approach the boundary values, $0$ or  $1$.
\begin{corollary}[$\mathcal{W}^*_E$ Under  $\rho_{ij}$ being 0 or 1] \label{Extrem} 
	Under general $\Sigma$, divide the study indices into $\mathcal{I}_1, \mathcal{I}_2$, where $\mathcal{I}_1= \{k : k \in \{1, \cdots, K-1 \}, \rho_{k K} = 0\}$ and $\mathcal{I}_2 = \{1,\cdots, K\} \setminus \mathcal{I}_1$.  For any $k \in \mathcal{I}_1$, if we know $\rho_{k k'} = 0$ for all $k' \in \mathcal{I}_2$,  $(\mathcal{W}^*_E)_k = 0$. In the special case, when $\rho_{kk'} \rightarrow 0, k,k' \in \{1, \cdots, K\}, i \neq j$, $\lambda_K = \gamma_K / \alpha^2_K $,  we have 
	\begin{align*}
		W^*_E \rightarrow \begin{pmatrix}
			0 \\ \vdots \\  1
		\end{pmatrix}
	\end{align*}
	When $\alpha_{1}^2 = \cdots = \alpha_{K}^2 = \alpha^2$, $\sigma_{1}^2 = \cdots = \sigma_{K}^2 = \sigma^2$, $\gamma_{1} = \cdots = \gamma_{K} = \gamma$, under $\rho_{k k'} \rightarrow 1$ for  $k, k' \in \{1, \cdots, K\}$ and  $k \neq k'$, the optimal weight and optimal estimation risk behave like the weight and risk in \citet{sheng2020}, with  $(\mathcal{W}^*_E)_1 = \cdots = (\mathcal{W}^*_E)_K$. Further, if $\Sigma = I_p$, $\lambda_1 = \cdots, \lambda_K = \lambda = {\gamma}/{\alpha^2}$, write $m(-\lambda) := m_{F_{\gamma}} (-\lambda)$, we have 
	\[(\mathcal{W}^*_E)_k = \frac{1}{K + (1 - K) \lambda m (-\lambda)}, k = 1,\cdots, K-1\]
	and 
	\[1^T \mathcal{W}^*_E = \frac{K}{K + (1 - K) \lambda m (-\lambda)} = 1 + \frac{(K-1) \lambda m (-\lambda)}{K + (1 - K) \lambda m (-\lambda)} > 1\]
\end{corollary} 

Corollary \ref{Extrem} states that when all correlations are zero, the limiting weights assigned to all source ridge estimators are zero, and the limiting weight assigned to the target ridge estimator is one. In this case the proposed transfer learning estimator becomes identical to a naive ridge estimator using only the target data, which should be the best that we can do when there is no relevant information in source populations. When all correlations are one, the limiting weights for all ridge estimators are the same, which is expected  since the information in all studies are essentially the same. Moreover, the sum of the weights is always larger than $1$. In this case the transfer learning estimator can be understood as a debiased estimator of the originally underestimated naive ridge estimators. This scenario is studied  in \citet{sheng2020} under the context of distributed learning. 

 Corollary \eqref{WR} demonstrates how the weight $\mathcal{W}^*_E$ changes as a function of $\rho$  under the setting $\Sigma=I$ and  $\rho_{k k'}=\rho$ for all $k$ and $k'$. We see the derivative of the weight ratio is always negative, implying  that the relative sizes of the weights assigned to source estimators  increase with respect to $\rho$. Since the ratio is at most $1$, Corollary \ref{WR} shows that the weight assigned to target estimator is never less than the weights assigned to source estimators.
 
\begin{corollary}[Monotone Change in Weight Ratio] \label{WR}
	Consider when $\Sigma = I_p$, $\alpha_{1}^2 = \cdots = \alpha_{K}^2 = \alpha^2$, $\sigma_{1}^2 = \cdots = \sigma_{K}^2 = \sigma^2$, $\gamma_{1} = \cdots = \gamma_{K} = \gamma$, $ \lambda_1 = \cdots, \lambda_K = \lambda = {\gamma}/{\alpha^2}$, $\rho_{k k'} = \rho$ for $k, k'= 1, \cdots, K; k \neq k'$. Again write $m(-\lambda) := m_{F_{\gamma}} (-\lambda)$, we have 
	\[(\mathcal{W}^*_E)_K = \frac{1/\rho + \frac{(1-\rho) (K-1) \{1 - \lambda m(-\lambda)\}}{  1 -  \rho \{1 - \lambda m(-\lambda)\}} }{[1-\rho \{1 - \lambda m(-\lambda)\}] +  K \{1 - \lambda m(-\lambda)\}}\]
	\[(\mathcal{W}^*_E)_{k} = \frac{1 + \frac{(\rho-1) [1 - \lambda m(-\lambda)]}{  1 -  \rho [1 - \lambda m(-\lambda)]} }{[1-\rho \{1 - \lambda m(-\lambda)\}] +  K \{1 - \lambda m(-\lambda)\}},\]
	\begin{align*}
		\frac{d \frac{(\mathcal{W}^*_E)_K}{(\mathcal{W}^*_E)_{k}}}{d \rho} = \frac{ - \Big[\{1 - \lambda m(-\lambda)\}\rho - 1\Big]^2 - \lambda m (-\lambda)}{\rho^2 \Big[\{1 - \lambda m(-\lambda)\} \rho \{1 - \lambda m(-\lambda)\} - 2\Big]^2}, 
\end{align*}
for $k = 1, \cdots, K-1$.
\end{corollary}

\begin{figure}
\includegraphics[height=5cm, width=0.4\textwidth]{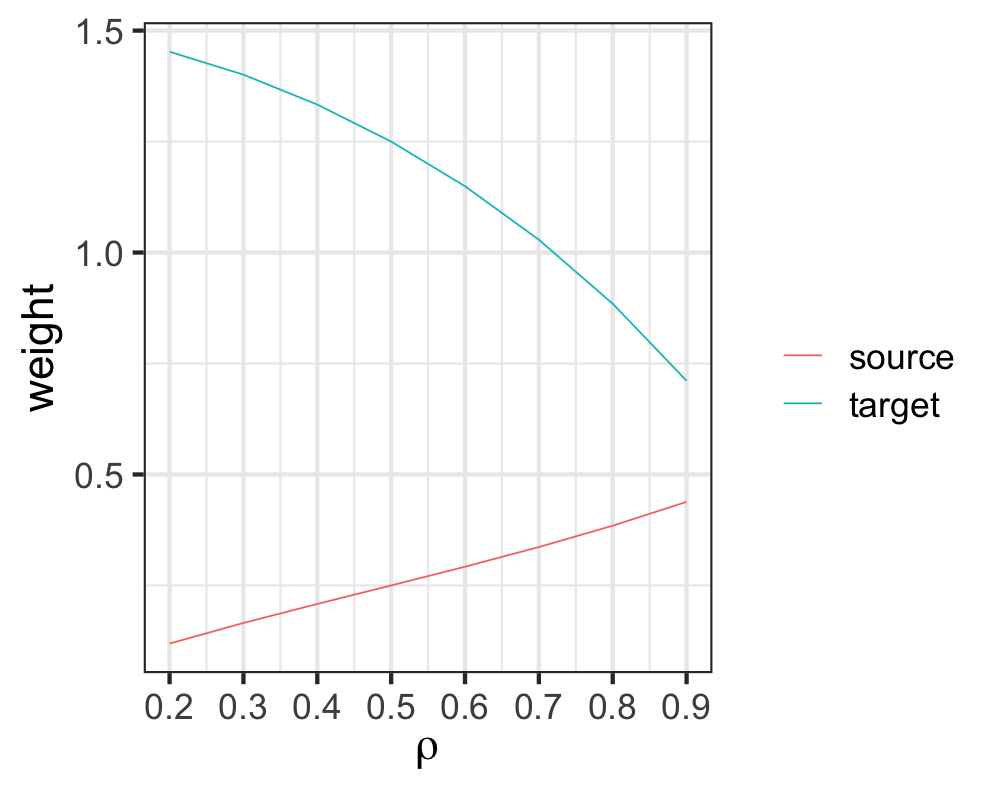}
\includegraphics[height=5cm, width=0.49\textwidth]{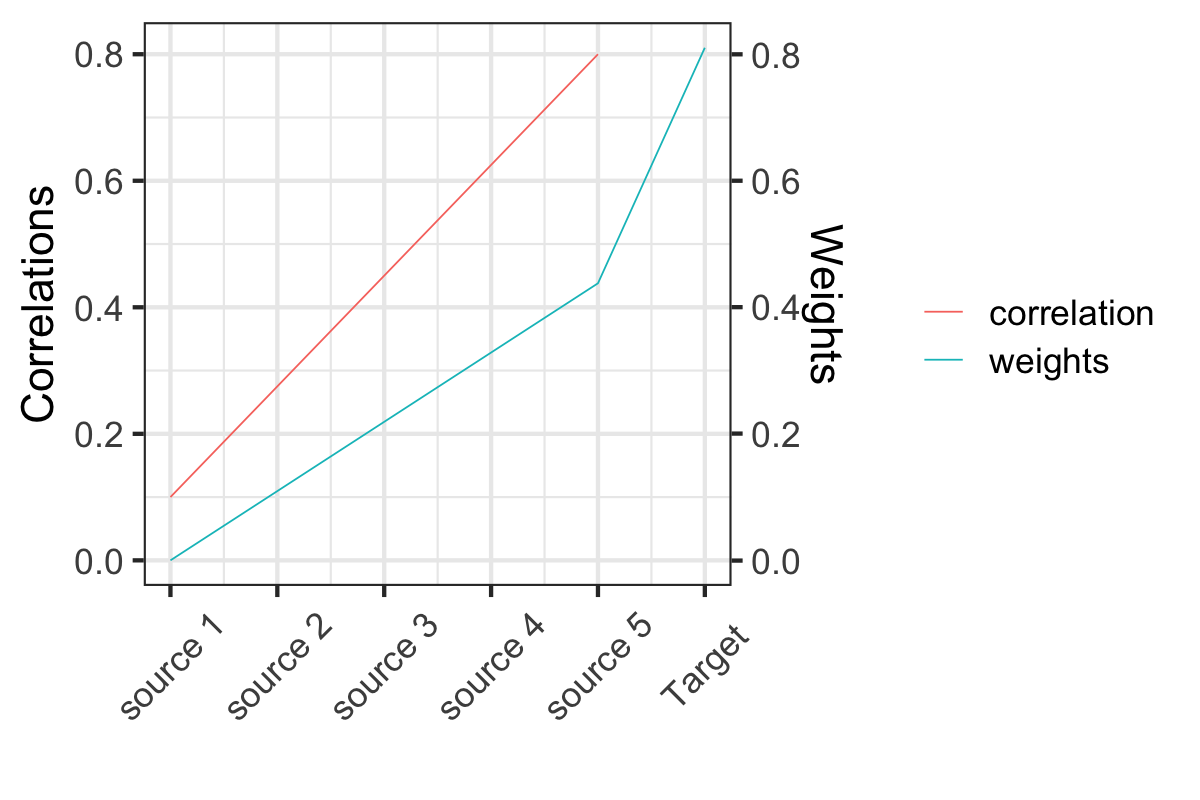}
\caption{Adaptivity of the weights as a function correlation of random coefficients. All six simulated studies have $100$ observations and $100$ predictors. Left: the correlations between studies are set to a common $\rho$. As the $\rho$ increases, more weights are shifted from the source studies to the target study. Right: each source study has  a different $\rho_{k K}, k = \{1, \cdots, K-1\}$, ranging  from $0.1$ to $0.8$. The plot shows that more weights are given to studies with higher $\rho_{k K}$, but all weights are less than $\mathcal{W}^*_K$
}
\label{weight_RhoNE_new}
\end{figure}

 We illustrate this adaptivity of the weight with respective to the correlation in Figure \ref{weight_RhoNE_new}. When the correlations are different across different populations, an explicit relationship between $\rho_{k K}, k = \{1, \cdots, K-1\}$  and the weights is hard to be found. However,  Figure \ref{weight_RhoNE_new} suggests that a similar adaptivity property still holds.

\section{Analysis of the estimation risk  $\mathcal{M}^*_K(\mathcal{W}_E^*)$ under $\Sigma = I_p$}

We next investigate the behavior of the limiting estimation risk using the optimal weight $\mathcal{M}^*_K(\mathcal{W}^*_E)$ under the setting when $\Sigma = I_p$. We demonstrate how the adaptivity of the weight $\mathcal{W}^*_E$  leads to improvement in estimation of $\mathcal{M}^*_K(\mathcal{W}^*_E)$ when $\rho$ changes.  For simplicity, we  assume $\rho_{k k'} = \rho$ for $k, k'= 1, \cdots, K$ and  $k \neq k'$,$ \lambda_1 = \cdots, \lambda_K = \lambda$, $\alpha_{1}^2 = \cdots = \alpha_{K}^2 = \alpha^2$, and  without loss of generality, $\sigma_{1}^2 = \cdots = \sigma_{K}^2 = 1$. Under these settings, Corollary \eqref{expEst} presents an explicit expression of  $\mathcal{M}^*_K(\mathcal{W}^*_E)$.
\begin{corollary}[Simple Expression for $\mathcal{M}^*_K(\mathcal{W}^*_E)$]\label{expEst}
	Consider when $\Sigma = I_p$, $\rho_{k k'} = \rho$ for $k, k'= 1, \cdots, K; k \neq k'$,$ \lambda_1 = \cdots, \lambda_K = \lambda$, $\alpha_{1}^2 = \cdots = \alpha_{K}^2 = \alpha^2$, $\sigma_{1}^2 = \cdots = \sigma_{K}^2 = 1$, $\gamma_{1} = \cdots = \gamma_{K} = \gamma$. Set $\lambda = {\gamma}/{\alpha^2}$ and write $m(-\lambda) := m_{F_{\gamma}} (-\lambda)$, we have 
		\begin{eqnarray*}
		&&\mathcal{M}^*_K(\mathcal{W}^*_E) =\alpha^2 - V^T (\mathcal{A} + \mathcal{R})^{-1} V = \frac{\alpha^2 [1 - \rho \{1 - \lambda m_{\gamma}(-\lambda)\}] }{ 1/ \rho  +  (K-1) \{1 - \lambda m_{\gamma}(-\lambda)\} } \\
		&+&  \frac{\alpha^2 (1 - \rho) (K-1) \{1 - \lambda m(-\lambda)\}\left\{1   -  \{1 - \lambda m(-\lambda)\} - \frac{1}{\rho(K-1)} + \frac{1}{(K-1)}\{1 - \lambda m(-\lambda)\}\right\} }{ \left[\frac{1}{\rho\{1 - \lambda m(-\lambda)\}}  +  (K-1)\right]
			\left[\frac{1}{\{1 - \lambda m(-\lambda)\}} + \rho\right]},
	\end{eqnarray*}
where
\[m(-\lambda) = \frac{-(1 - \gamma + \lambda) + \sqrt{(1 - \gamma + \lambda)^2 + 4 \gamma \lambda}}{2 \gamma \lambda}.\]
\end{corollary}

We  observe that the Trans-Ridge estimation risk  reduces to the estimation risk of a naive ridge regression fitted with only the target data when $\rho \rightarrow 0$, and it reduces to the estimation risk of the distributed learning ridge \citet{sheng2020} when $\rho \rightarrow 1$. When $\lambda$ is optimal for each individual ridge estimator ($\lambda = \gamma / \alpha^2$), the estimation risk for each ridge estimator is $\gamma m (-\lambda)$, and the term $1 - \lambda m(-\lambda)$ is the amount of improvement from the null estimator $\beta_{null} = 0$ divided by the \textsc{snr} $\alpha^2$. When this quantity, which only depends on $\gamma, \alpha^2$, is known,  one can compute the dynamic of $\mathcal{M}^*_K(\mathcal{W}^*_E)$ with respect to $K$ and $\rho$. In the Supplemental Material, we show how to find suitable choices for $K$ so $\mathcal{M}^*_K(\mathcal{W}^*_E)$ strictly decreases with $\rho$. 

 Corollary \eqref{expEst} leads to an upper bound on the limiting estimation risk under the  optimal limiting weights:
 \begin{align*}
 	\mathcal{M}^*_K(\mathcal{W}_E^*) &\leq \alpha^2  -  \frac{1}{1 + a^* + \frac{c_2}{c_1}}\frac{\alpha^4}{c_1} \left( \sum_{i = 1}^{K-1} \rho_{i K}^2 + 1 \right) \{1 - \lambda m (-\lambda)\}^2.
 \end{align*}
where 
\begin{align*}
	c_1 &:= \alpha^2 \{1 - 2 \lambda m (-\lambda) + \lambda^2 m(-\lambda)^2\}\\
	c_2 &:= \lambda^2 \alpha^2  \{m'(-\lambda) - m_(-\lambda)^2\} + \gamma \{m (-\lambda) - \lambda m'(-\lambda)\} \\
	a^* &:= \sup_{i \in \{1, \cdots, K\}} \sum_{k = 1, k \neq i}^{K} |\rho_{i k}| 
\end{align*}
If further $\rho_{k k'} = \rho$ for $k, k'= 1, \cdots, K$ and  $k \neq k'$, the same bound holds with $a^* = (K-1)\rho$.

\begin{figure}[h]
\centering
\includegraphics[height=5cm, width=0.8\textwidth]{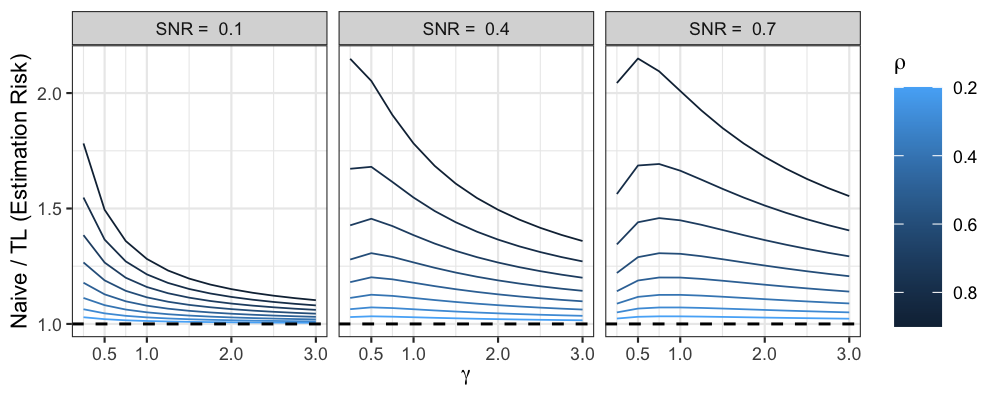}
\caption{Ratio of the estimation risk of Trans-Ridge over one sample ridge regression as a function of the strength of correlation $\rho_{iK}$, denoted by the colors, and the aspect ratio $\gamma= p / n$ as $n,p \rightarrow \infty$.}
\label{EstCompare}
\end{figure}

  A simple analysis of the upper bound reveals that it strictly decreases with every $\rho_{iK}$ for $i \in \{1, \cdots, K-1\}$ under very mild conditions on the magnitude of $\rho_{i K}$ (see Supplemental materials). The upper bound arises when controlling the quadratic term $a^T \Sigma_\beta a$ where $\Sigma_\beta$ is the covariance matrix of linear coefficients in each population, and the eigenvalues of $\Sigma_\beta$ is replaced by the largest one. $a^*$ is from controlling the magnitude of the largest eigenvalues of $\Sigma_\beta$. 
To demonstrate this point, 
Figure \ref{EstCompare} shows a plot of the relationships between $\mathcal{M}^*_K(\mathcal{W}^*_E)$ and $\rho_{iK}$. We define the liming risk of a naive ridge estimator under optimal tuning as $\mathcal{M}^*_0$, and plot the ratio of $\mathcal{M}^*_0$ and $\mathcal{M}^*_K(\mathcal{W}^*_E)$ on the y-axis, where  individual ridge estimators are not affected by $\rho_{ij}$. We later  show that similar patterns hold for prediction risks. From the same figure, we also find that if \textsc{snr} is small, the improvement brought by Trans-Ridge estimator monotonically increases when $\gamma$ gets larger. This monotone relationship does not hold when \textsc{snr} is large enough, and a bump would appear on the curve. 

\section{Weight Estimation Based on Prediction Risk}
\subsection{Optimal Prediction Weight}
 An alternative to estimate the weight vector is by  minimizing the expected risk in prediction.  We first define the expected prediction risk for a Trans-Ridge estimator
\begin{definition}\label{predR}
	The expected prediction risk of a Trans-Ridge estimator with weight $W$ is defined as 
	\begin{align*}
		r(W) = E_{\epsilon_{1}, \cdots, \epsilon_K, x_0}\left\{(y_0 - \sum_{k = 1}^{K} W_k \hat{\beta}_{k} x_0^T )^2 
		| X_1, \cdots, X_K\right\}
	\end{align*}
where $(x_0, y_0)$ is an independent random test example drawn from target population. The expectation is taken over the randomness in both $x_0$ and error terms in each population conditioned on all design matrices.
\end{definition}

Similar to the estimation risk, we can find a weight that minimizes the expected prediction risk, and  denote this optimal prediction weight  as $W_P^*$. Using $E_{\epsilon}$ to denote the expectation over $\epsilon_1, \cdots, \epsilon_K$,  the following theorem gives the finite-sample expressions for optimal prediction weight and the corresponding prediction risk.
\begin{theorem}[Finite sample optimal weight for prediction]\label{finOptPred}
	Define the following matrices
	\begin{align*}
		B &:= \begin{pmatrix}
			E(\hat{\beta}_1) & \cdots & E(\hat{\beta}_K)
		\end{pmatrix} = \begin{pmatrix}
			Q_1 \beta_1 & \cdots & Q_K \beta_K
		\end{pmatrix} \\
		D &:= B^T \Sigma \beta_K = vec[\beta_k Q_k \Sigma \beta_K] \\
		C &:=  mat[E_{\epsilon}(\hat{\beta}_k^T \Sigma \hat{\beta}_{k'})] \\
		F &:= diag[\sigma_{k}^2 \tr\{(\hat{\Sigma}_k + \lambda_{k} I_p)^{-1} \hat{\Sigma} (\hat{\Sigma}_k + \lambda_{k} I_p)^{-1} \Sigma\} / n_k]
	\end{align*}
Then the optimal weight is given by 
\[W_P^* = (C + F)^{-1} D, \]
and the correspnding predictin risk is given by
\[r (W_P^*) = \sigma_{K}^2 +  \beta_K^T \Sigma \beta_K -  D^T (C + F)^{-1} D.  \]
\end{theorem}

The limits of $D, F$ and diagonal entries of $C$  can be expressed in empirical terms with the standard techniques \citep{dobriban2018high}. The off-diagonal entries of $C$, like $A_{ij}$ in Theorem \ref{optWeight}, is an interplay between the spectrum of $\hat{\Sigma}_i$ and $\hat{\Sigma}_j$. Further, since the prediction risk is over the randomness of $x_0$, it also involves the spectrum of population covariance matrix $\Sigma$ through 
$P_{k k'}  := \tr\{\Sigma (\hat{\Sigma}_k + \lambda_k)^{-1} (\hat{\Sigma}_{k'} + \lambda_{k'})^{-1} \}/ p.$
The following Lemma, similar to  Lemma \ref{aijasp},  expresses the limit of $P_{k k'}$ in sample quantities under different assumptions.
\begin{lemma}\label{predijasp}
	With assumption  \ref{aspRMT}, and under $n_k, p \rightarrow \infty$, $p / n_k \rightarrow \gamma_k$,  we have $P_{k k'} \rightarrow_{a.s.} \mathcal{P}_{k k'}$. The expressions for $\mathcal{P}_{k k'}$ under different assumptions are summarized below. We use a shorter notation $m_k(-\lambda_k) := m_{F_{\gamma_k}} (-\lambda_k), m'_k(-\lambda_k) = m'_{F_{\gamma_k}} (-\lambda_k)$
	\begin{enumerate}
	\item Assume $n_1 = \cdots = n_K = n$ so $\gamma_1 = \cdots = \gamma_K = \gamma$ and use $\lambda_1 = \cdots = \lambda_K = \lambda$. Write $m(-\lambda) := m_{F_{\gamma}} (-\lambda), m'(-\lambda) = m'_{F_{\gamma}} (-\lambda)$ we have for $k \neq k'; k, k'\in \{1, \cdots, K\}$
	\[\mathcal{P}_{k k'} =  \frac{m(-\lambda) - \lambda m'(-\lambda)}{1 -  \gamma +  \gamma \lambda^2 m'(-\lambda)}. \]
	\item Under the assumption $\Sigma = I_p$ and assumption \ref{assmoment},
	\[\mathcal{P}_{k k'} =   m_{k} (-\lambda_k)  m_{k'} (-\lambda_{k'}). \]
	\item Under Assumptions \ref{assmoment} and  \ref{assIso}, 
	\[\mathcal{P}_{k k'} = \frac{\lambda_k m_{k} (-\lambda_k) - \lambda_{k'} m_{k'} (-\lambda_{k'})}{\lambda_{k} \lambda_{k'} (m_{k'} (-\lambda_{k'}) - m_{k} (-\lambda_k))}. \]
	\end{enumerate}
\end{lemma}

The key to obtain the limiting expression for $\mathcal{P}_{k k'}$  lies in breaking the integral into pieces of functions of $m_{F_{\gamma_{K}}}(-\lambda_{k})$. Applying this Lemma  leads to the limiting form of prediction risk and optimal prediction weight given in  Theorem \ref{predOptAsp}.
\begin{theorem}[Asymptotic Optimal Prediction Weights and Prediction Risk] \label{predOptAsp} Assume  assumptions \ref{aspRCR},\ref{aspTL} and the  \ref{aspRMT} hold. We  further assume  that the entries of $Z$ have a finite $(8 + \epsilon)$-th moment for some $\epsilon > 0$, and all eigenvalues of $\Sigma$ are uniformly bounded away from zero and infinity. Assume  that variables are scaled so $\Sigma_{ii} = 1$ for $i = 1, \cdots, p$. We have 
	\begin{enumerate}
		\item $D \rightarrow_{a.s.} \mathcal{D}$, where for $k = 1, \cdots K-1$,
		\[\mathcal{D}_k = \rho_{k K} \sigma_k \sigma_K \alpha_k \alpha_K \left[1 - \frac{\lambda_{k}}{\gamma_k} \left\{\frac{1}{\lambda_{k} v_{F_{\gamma_k}} (-\lambda_k)}\right\} \right], \] 
	and 
		\[\mathcal{D}_K = \sigma_K^2 \alpha_K^2 \left[1 - \frac{\lambda_{k}}{\gamma_K} \left\{\frac{1}{\lambda_{K} v_{F_{\gamma_K}} (-\lambda_K)}\right\}\right].\]
		\item $C \rightarrow_{a.s.} \mathcal{C}$, where for $k = 1, \cdots, K$,
		\begin{align*}
			\mathcal{C}_{kk} &= \sigma_k^2  \alpha_k^2 \left[ 1 - 2 \frac{\lambda_k}{\gamma_k} \left\{\frac{1}{\lambda_{k} v_{F_{\gamma_k}} (-\lambda_k)}\right\} +  \frac{\lambda_k^2}{\gamma_k} \frac{v_{F_{\gamma_k}} (-\lambda_k) - \lambda_k v'_{F_{\gamma_k}} (-\lambda_k)}{[\lambda_k v_{F_{\gamma_k}} (-\lambda_k)]^2}\right], 
		\end{align*}
and 	for $k, k' = 1, \cdots, K$ and $k \neq k'$,  
	\begin{align*}
		\mathcal{C}_{k k'} &= \rho_{k  k'} \alpha_k \sigma_k \alpha_{k'} \sigma_{k'}  \left[1 - \frac{\lambda_k}{\gamma_k} \left\{\frac{1}{\lambda_{k} v_{F_{\gamma_k}} (-\lambda_k)}\right\} - \frac{\lambda_{k'}}{\gamma_{k'}} \left\{\frac{1}{\lambda_{k'} v_{F_{\gamma_{k'}}} (-\lambda_{k'})}\right\} + \lambda_k \lambda_{k'} \mathcal{P}_{k k'}. \right]
	\end{align*}
		\item  $F$ is a diagonal matrix, and $F \rightarrow_{a.s.} \mathcal{F}$, where 
		\begin{align*}
			\mathcal{F}_{kk} = \sigma_k^2 \gamma_k \left[\frac{1}{\gamma_k} \left\{\frac{1}{\lambda_k v_{F_{\gamma_k}}(-\lambda_k)} - 1\right\} - \lambda_k \frac{1}{\gamma_k} \frac{v_{F_{\gamma_k}}(-\lambda_k) - \lambda_k v'_{F_{\gamma_k}}(-\lambda_k)}{[\lambda_k v_{F_{\gamma_k}}(-\lambda_k)]^2}\right].
		\end{align*}
	\end{enumerate}
	The limiting optimal prediction weight is
$W_P^* \rightarrow_{a.s.} \mathcal{W}_P^* =(\mathcal{C} + \mathcal{F})^{-1} \mathcal{D}$ and the corresponding 
 limiting estimation risk under optimal weights is 
$r(W_P^*) \rightarrow_{a.s.} \scripty{r}(\mathcal{W}_P^*) = \sigma_{K}^2 +  \alpha_{K}^2 \sigma_{K}^2 -  \mathcal{D}^T (\mathcal{C} + \mathcal{F})^{-1} \mathcal{D}.$
\end{theorem}
\begin{figure}[h]
     \centering
     \begin{subfigure}[b]{0.6\textwidth}
         \centering
         \includegraphics[width=\textwidth]{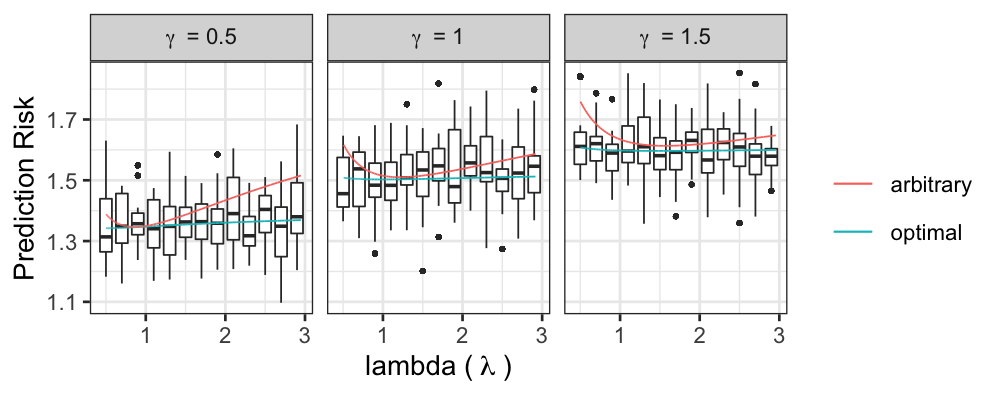}
     \end{subfigure}
        \\[\smallskipamount]
     \begin{subfigure}[b]{0.6\textwidth}
         \centering
         \includegraphics[width=\textwidth]{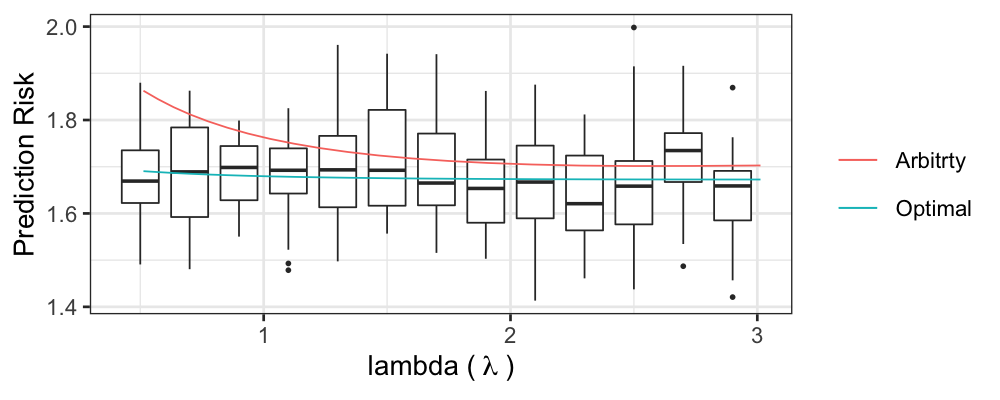}
     \end{subfigure}
     \\
     \begin{subfigure}[b]{0.6\textwidth}
         \centering
         \includegraphics[width=\textwidth]{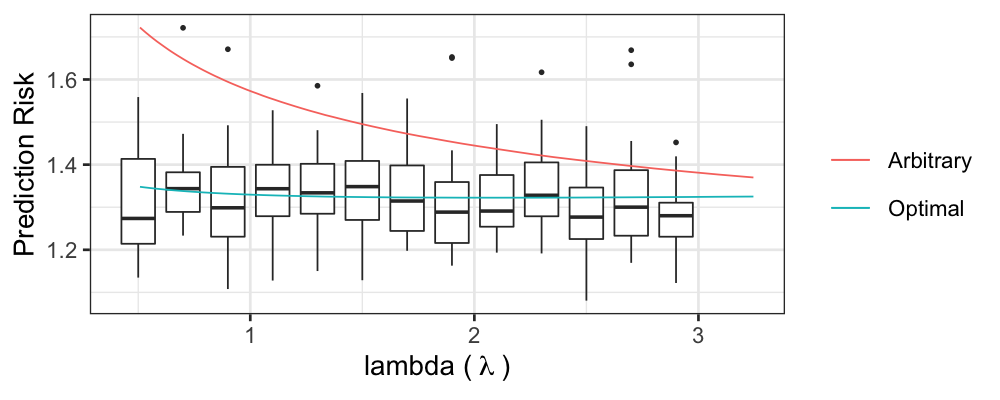}
         \label{fig:five over x}
     \end{subfigure}
        \caption{Comparison between theoretical prediction risk and empirical prediction risks. Top: all data sources have the same sample size. It seems that an optimally-tuned trans-ridge with arbitrary weights can achieve similar performance as the trans-ridge with optimal weights.
		Middle: The second scenario where each study have different sample sizes under identity population covariance matrix.
		Bottom: The same setup as the middle plot, but with a Toeplitz population covariance matrix. Using the optimal weights brings more improvment in prediction risk compared to the identity covariance matrix.}
		\label{PredTheo_Sigma}
\end{figure}

We note that the mild assumption $\Sigma_{ii} = 1$ allows us to easily compute the $\tr (\Sigma) / p$. In Figure \ref{PredTheo_Sigma}, we again show three different plots,  each corresponding  to a scenario in Lemma \ref{predijasp}.  The simulation setups for these three plots are exactly the same as the setups of the estimation risk (Figure \ref{EstTheo_Sigma}). The plots show that the theoretical prediction risks   agree well with the empirical prediction risks in all scenarios. 
\begin{figure}[h]
\centering
\includegraphics[height=5cm, width=0.8\textwidth]{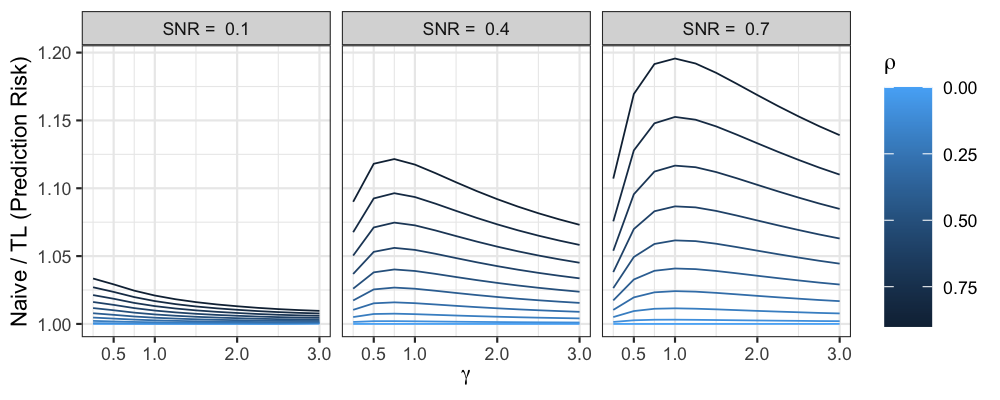}
\caption{Ratio of prediction risk of Trans-Ridge over one sample ridge regression as a function of the strength of correlation $\rho_{iK}$  denoted by the colors, and the aspect ratio $\gamma = p / n$. }
	\label{PredCompare}
\end{figure}
Figure \ref{PredCompare} shows the plots of the ratio of the prediction risk from Trans-Ridge and single sample ridge regression. First,  the improvement on the prediction risk is not as large as the improvement of the estimation risk.  Even when the study correlation is as high as $0.9$, the magnitude of improvement is at most  $20\%$. This may suggest dominant part of prediction risk is in fact irreducible. Secondly, just like for estimation risk, when no source study is useful ($\rho = 0$), the risk of Trans-Ridge estimator is the same as the naive ridge estimator. A similar result to Corollary \ref{Extrem}  for optimal prediction weights can be easily derived as the expressions of $\mathcal{W}_P^*$ and $\mathcal{W}_E^*$ share the same form. Thirdly,  when$\rho$ increases, the improvement from Trans-Ridge estimator seems to strictly increase. The rate of this increase is higher for larger $\rho$.
Lastly,  the regime of $\gamma$ where Trans-Ridge estimator leads the largest improvement seems to increase with \textsc{snr}. When \textsc{snr} is  large, the improvement appears to be a  concave function of $\gamma$. The prediction risk of optimally tuned naive ridge regression is a strictly increasing function of $\gamma$. This argument combined suggest there may be a regime of $\gamma$ where the prediction risk of Trans-Ridge estimator decreases with $\gamma$. Although the amount of improvement is lower for large $\gamma$, the risk curves seems to flatten as well, indicating smaller prediction risk even for large $\gamma$. 

\section{Application to real data sets}
We demonstrate the proposed transfer learning  ridge estimator using  Penn medicine biobank, a large biobank of genomic data linked to electronic health record phenotype data. Three genetically linked lipids phenotypes generally predictable by \textsc{snp}s are considered,  including high density lipoprotein cholesterol (\textsc{hdl}), low density lipoprotein cholesterol (\textsc{ldl}) and triglycerides (\textsc{tri}). We apply the methods  in two common scenarios where source studies are available. The first scenario is when data for the same phenotype are available from different studies, however researchers are only interested in the polygenic scores of a particular (target) study. We study this scenario by dividing approximately $12000$ subjects in the data set into $4$ equal-sized datasets, and one of the dataset is chosen to represent the target population. The predictors are chosen from $650281$ \textsc{snp}s by univariate thresholding. We calculate the univariate effect of each \textsc{snp}s with only the target data, and select the top $3000$ \textsc{snp}s as predictors. The genetic heritability/correlation are estimated by the bivariate GREML method \citep{lee2012estimation}.

We demonstrate the prediction results using both the optimal estimation weight and the optimal prediction weight for Trans-Ridge. Three competing methods are considered; namely, naive ridge estimator fitted on the target population, the Trans-Lasso regression \citep{li2021transfer} and the null predictor using sample mean. The penalization parameter is assumed to be the same for each study for all methods and is tuned by cross-validation. The candidates $\lambda$ for Trans-Ridge are multiples of the optimal naive ridge penalization parameter $\gamma / \alpha^2$, and the candidates $\lambda$ for Trans-Lasso are multiples of  $2 \log p / n$. Here $p$ is the number of \textsc{snp}s and $n$ is the number of training observations. We show below the testing mean square errors evaluated on $1000$ held-out subjects for 5 methods.
\begin{table} 
\begin{center}
	\caption{Analysis of lipid traits data. The table entry is the testing data mean square error, where each trait is a target, and other two sets of the same trait are source traits. }
	\label{Tstudy}
		\begin{tabular}{ c ccccc}
			& OptEst & OptPred & Naive & Lasso & Null \\
			\textsc{hdl} & 0.792 & 0.682 & 0.744 &  0.881 & 0.888 \\
			\textsc{ldl} & 0.884 & 0.805 &  0.853 & 1.018 & 1.032 \\
			\textsc{tri} & 0.931 & 0.920 & 0.936 & 0.965 & 0.963\\
		\end{tabular}
\end{center}
		\begin{tablenotes}
			\item[1] OptEst: Trans-Ridge with estimation-based weights; OptPred: Trans-Ridge with prediction -based weights;
			Naive: ridge regression using only target data; Lasso: Trans-Lasso; Null: sample mean. 
		\end{tablenotes}
\end{table}
From Table \ref{Tstudy},  we see Trans-Ridge clearly dominates the other methods in this setting, where information is indeed available in the source populations. Trans-Ridge with the optimal prediction weights gives the best prediction performance  for all three traits. We also observe that even when the naive ridge estimator is worse than null predictor (\textsc{tri}), that is nothing useful could be learned from  only the data of target population, by adding such coefficient estimates using the  optimal weights, we still improve  the prediction performance. 

As the second set up, we consider when source datasets of related traits are available for the target trait. We use each of three traits as target trait and the rest two as the source traits. We again divide the approximately twelve thousand observations into three groups, one for each trait. We use the same set of  \textsc{snp}s selected using univariate screening as the first set up. The penalization parameters are selected in the same way, and the held-out test mean-square errors are shown below.

\begin{table} 
	\begin{center}
			\caption{Analysis of lipid trait data set. The table entry is the testing data mean-square error, where each trait is a target, and the other two traits are source traits.  }
		\label{Ttrait}
		\begin{tabular}{  c ccccc}
			& OptEst & OptPred & Naive & Lasso & Null \\
		\textsc{hdl} & 0.880 & 0.875 & 0.950 &  1.021 & 0.907 \\
		\textsc{ldl} & 0.903 & 0.914 &  0.915 & 0.982 & 1.003 \\
		\textsc{tri} & 0.908 & 0.912 & 0.912 & 0.951 & 0.982\\
		\end{tabular}
	\end{center}
\begin{tablenotes}
	\item[1] OptEst: Trans-Ridge with estimation-based weights; OptPred: Trans-Ridge with prediction-based weights;
Naive: ridge regression using only target data; Lasso: Trans-Lasso; Null: sample mean. 
\end{tablenotes}
\end{table}

In this scenario, we notice that the performance gain from Trans-Ridge is not as large as the first set up. This is expected, as the useful information is much less across different genetic traits; the genetic correlation estimated across the three traits are well below $0.5$ compared to $\sim 1$ for the same trait. Although Trans-Ridge still shows dominance, the optimal estimation weights show better prediction  performance than the optimal prediction weights. We attribute this to the robustness of  estimation of the weights.
\section{Discussion}
We have presented  transfer learning methods based on random coefficient ridge regressions, where the target model coefficient is estimated as a weighted combination of ridge estimate of each study. We develop two computationally efficient methods for estimating the weights by minimizing the  estimation risk or  prediction risk. In general, if the goal is prediction, we suggest to use the weight estimated by minimizing the prediction risk.  Although in theory, this should give small prediction errors, in real data analysis when the sample sizes are not too large, weights estimated by minimizing the estimation risk may sometimes give better prediction performance. 

We have derived  the limiting predicting risk for the target model by using random matrix theorem under general covariance structure of the predictors. As a comparison,  \citet{sheng2020} provides the limiting prediction risk only when population covariance matrix is the identity matrix. In addition, if the sample sizes $n_k$ or the  regularization  $\lambda_{k}$  parameters are different across $K$ populations, the limiting estimation risk in \citet{sheng2020} cannot be easily evaluated  for general covariance matrix $\Sigma$. \citet{zhao2019cross} provides the asymptotic out-of-sample $R^2$ of for regular single sample ridge estimator. However, they did not study the  prediction risk as defined in \eqref{predR}. 

\section*{Acknowledgments}
This research was supported by NIH grants. We would like to thank Dr. Jiaoyang Huang for  discussions on random matrix theorems in the derivations.

\section*{Supplementary material} Supplementary material available online includes proofs of all the lemmas and theorems. 

\bibliographystyle{unsrtnat}
\bibliography{references}  






\end{document}